\documentclass[twocolumn]{svjour3}          
\smartqed  
\usepackage{graphicx}
\usepackage{mathptmx}      

\begin{document}

\title{The Robot Economy: Here It Comes
}


\author{Miguel Arduengo        \and
        Luis Sentis 
}


\institute{Miguel Arduengo \at
              Human Centered Robotics Lab (University of Texas at Austin) \\
              \email{miguel.arduengo@upc.edu}           
           \and
           Luis Sentis \at
           Human Centered Robotics Lab (University of Texas at Austin)\\ 
           Department of Aerospace Engineering (University of Texas at Austin)\\
           \email{lsentis@austin.utexas.edu}
}

\date{}

\maketitle

\begin{abstract}
Automation is not a new phenomenon, and questions about its effects have long followed its advances. More than a half-century ago, US President Lyndon B. Johnson established a national commission to examine the impact of technology on the economy, declaring that automation ``can be the ally of our prosperity if we will just look ahead''. In this paper, our premise is that we are at a technological inflection point in which robots are developing the capacity to greatly increase their cognitive and  physical capabilities, and thus raising questions on labor dynamics. With increasing levels of autonomy and human-robot interaction, intelligent robots could soon accomplish new human-like capabilities such as engaging into social activities. Therefore, an increase in automation and autonomy brings the question of robots directly participating in some economic activities as autonomous agents. In this paper, a technological framework describing a robot economy is outlined and the challenges it might represent in the current socio-economic scenario are pondered.
\end{abstract}
\section{Introduction}

\begin{figure}
    \centering
        {\includegraphics[width=1.0\linewidth]{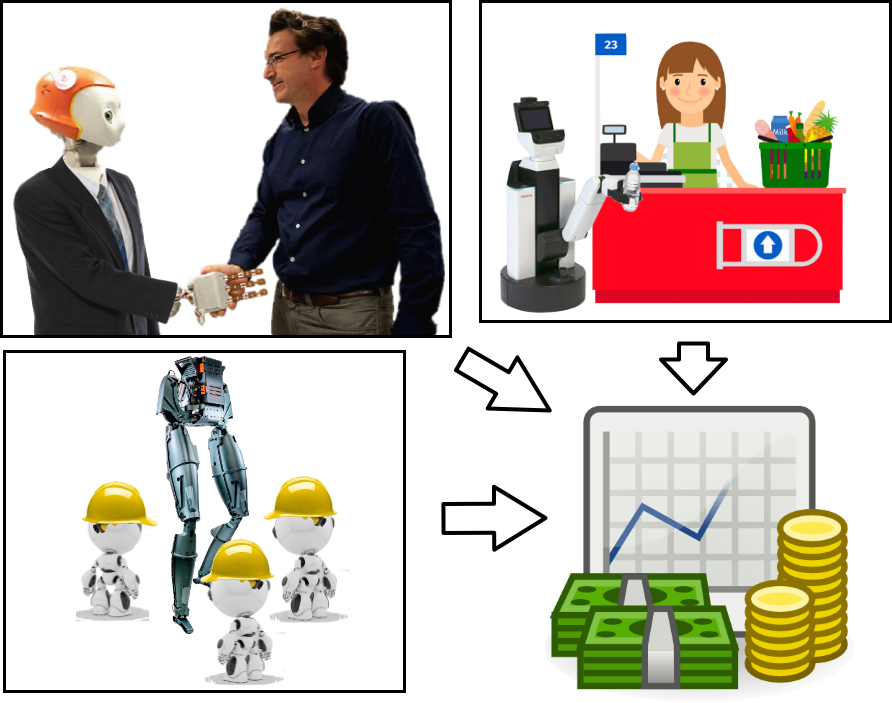}}
\caption{Robots are rapidly developing capabilities that could one day allow them to participate as autonomous agents in economic activities with the potential to change the current socio-economic scenario. Some interesting examples of such activities could eventually involve engaging into agreements with human counterparts, the purchase of goods and services and the participation in highly unstructured production processes.}
\label{Fig1}
\end{figure}

A robot, for our purposes, is a reprogrammable machine designed to execute diverse physical tasks. Embodied robots are every day acquiring more ``human-like'' capabilities such as dexterous manipulation and mobility in unstructured and dynamic environments.  For robots to be intelligent they would need to perceive, adapt, interact, move around, and manipulate objects—or a reasonable subset of these actions—in similar ways than humans do.

Disembodied Artificial intelligent systems such as stock market exchanging bots take decisions and adapt to new situations based on formal specifications provided by humans. In a vastly more sophisticated world, AI systems will also need physical embodiment to directly experience the natural world and provide non-trivial behavior capabilities without direct human intervention \cite{Kuipers2006}.
Advances in artificial intelligence are enabling greater autonomy for robots to make decisions in open worlds and unstructured environments. With increasing levels of autonomy and human-robot interaction, intelligent robots could soon accomplish new human-like capabilities such as engaging in economic agreements involving the exchange and consumption of services and goods with humans or other robot counterpart (Figure \ref{Fig1}). 

A robot economy (``robonomics'' \cite{Crews2016,Danilov2018,Ivanov2017a,AIRA2018}) would be an economic system in which intelligent robots act as autonomous agents with the capacity to replicate some human behaviors in various key economic activities.
The participation of robots in the economy has, so far, only taken place effectively for the production of goods. As a result, most of the published studies have focused on the impact of automation on economic growth, employment and income distribution \cite{Acemoglu2018b,Acemoglu2018a,Sachs2015,Berg2016,Berg2018,Brynjolfsson2014,Petropoulos2018,Ford2015,Freeman2015,Vermeulen2018}.
As noted by \cite{Acemoglu2017}, technological innovations can affect employment in two ways:
\begin{itemize}
\item By directly displacing workers from tasks they were previously trained for (displacement effect).
\item By increasing the demand for labor in new industries or jobs that arise or develop as a result of technological progress (productivity effect).
\end{itemize}
Therefore, to analyze the current impact of intelligent robotics, the main question is which of the two effects, displacement or productivity, could arise \cite{Petropoulos2018}.
Broadly speaking, two narratives have emerged \cite{Berg2016}:
\begin{itemize}
\item Technology pessimists, who consider that robonomics would evolve towards an economy with great inequality and class conflicts \cite{Crews2016,Sachs2015,Berg2018,Freeman2015,Sachs2012}.
\item Technology optimists, who point out that income growth raises the demand for labor in sectors that produce non-automatable goods or services
\cite{Acemoglu2018b,Acemoglu2018a,Acemoglu2017,Autor2014,Dellot2017}.
\end{itemize}
On the other hand, it should be considered that new technologies
depend on human and equipment capital, labor, and the role of institutions. And therefore, their impact is subject to various macroeconomic factors \cite{UNCTAD2017}. Thus, complex issues, such as the proliferation of low income jobs and ``zero-sum'' economic activities \cite{Turner2018}, need to be contemplated.

The debate between pessimists and optimists is unsettled, although there seems to be a certain consensus in accepting that a robot economy will inevitably have a redistribution effect that, to a large extent, will be positively regulated by our current institutions.

The starting point for this paper is that we are effectively in a technological inflection point in which intelligent robots are rapidly enabling the ability to perform cognitive and physical work, and perhaps become participants in a whole set of new economic activities. In the following sections: First, we develop a brief description of the essential characteristics of a robot economy (section 2). Second, we analyze a framework for which a robot economy could arise (section 3). Third, we consider the foreseeable impact of the robot economy in the current socio-economic environment (section 4). Fourth, these impacts are analyzed by means of a simple robotic model (section 5). Finally, in section 6, we briefly discuss robot economy challenges, such as regulatory policy, ethics or law.

\section{The Robot Economy}

A robot economy is a scenario in which intelligent robots would produce and provide many goods and services and also participate as autonomous agents in the exchange markets \cite{Ivanov2017a}. In a robot economy, intelligent robots can perform economic operations autonomously. For such activities to happen, robots must have the opportunity to create and undertake digital contracts for their services or operations, so that they can be fully integrated as autonomous agents into the human economy.

The economy is usually defined as the set of services and means for satisfying human needs in our societies through available resources. We consider that an essential issue for future economies will be to guarantee the satisfaction of all basic human needs. Therefore, in the value chain of any economic activity the final outcome must try to achieve this essential principle. This might result, among other things, in that although robots will be able to act as autonomous agents, they will not be able to obtain ownership over available resources. Therefore, robots must always perform tasks on a contractual basis in which they act as intermediaries in the realization of activities whose final result is to satisfy human basic needs.

Thus, in a safe and human-dependent robot economy we are contemplating three basic rules \cite{AIRA2018}:
\begin{enumerate}
\item A robot economy has to be developed within the framework of
the digital economy.
\item The economy of robots must have internal capital that can support the market and reflect the value of the participation of robots in our society.
\item There seems to be no justification for robots to have property rights and therefore they should operate only on the basis of contractual responsibilities.
\end{enumerate}
 
During economic-financial transactions, robots can adopt both the role of the agent that originates the transaction (supplier, seller) and the role of the agent receiving the transaction (buyer). On the supply side, some aspects of a robot economy are already developed, since industrial robots have been incorporated significantly in manufacturing in numerous sectors. Robots improve productivity when used for tasks they perform more efficiently and with higher quality than humans. It is expected that advances in artificial intelligence and machine learning will increase further the number of tasks that can be automated \cite{Acemoglu2018b}. 

On the demand side, a robot economy has not achieved any significant degree of development yet. However, future robots are expected to be able to purchase products and services on behalf of their owners. This means that they will affect consumer behavior \cite{Ivanov2017b}. In addition, robots participating as autonomous agents in the exchange of goods and services with humans or other robots, could be considered buyers themselves. Thus, from a demand perspective, the boundaries between robots and humans as consumers could be somewhat blurred
\cite{Ivanov2017b,Scheutz2009}.

In summary, recent research advancements in robotics and AI algorithms point to the development of some aspects of a robot economy with more advances to come soon.

\section{Technological framework for the robot economy}

For intelligent robots to participate in a robot economy, we highlight several technology requirements:
\begin{enumerate}
\item The need to autonomously perform tasks related to particular economic activities and communicating to other entities about the status and  execution of these tasks in accordance with the terms of a digital contract.
\item The need for real-time communications that allow prompt interactions with humans and other robots.
\item The need to formalize contractual liabilities through ``smart
contracts'', digital agreements that can incorporate complex contractual relationships, and that are self-executing
(zero ambiguity) and self-verifying (hard guarantees) \cite{Cardenas2018}. That is, neither the will of the parties to comply with their word nor the dependency on a third party (i.e. a legal system) are required \cite{Song2018}.
\item The need to carry out financial transactions using digital media. 
\end{enumerate}

In the following subsections we describe a framework that could meet these requirements with current technology and that would allow, therefore, building foundations for a potential robot economy.

\subsection{Robot Operating System (ROS)}

Intelligent robots require adequate tools for programming their tasks regarding economic activity, which could be enabled by the ``Robot Operating System (ROS)''. ROS is a Linux-based open-source robotic middle-ware which works with a publisher/subscriber model and nowadays it is acknowledged to be the standard software for larger parts of the worldwide robotics community \cite{Kehoe2015}. In the last years, ROS is gaining pace, extending to many platforms. For example, the Java ROS library, called rosjava, allows Android applications to be developed for robots \cite{ROSAndroid2019}. Also, Microsoft has just released a version of ROS for Windows platforms \cite{ROSWindows2019} and the ROS Industrial project is rapidly developing extending the advanced capabilities of ROS to the manufacturing sector \cite{ROSIndustrial2019}. Moreover, the ROS open-source software provides cloud support to build robotic applications in a decentralized network \cite{Toffetti2019}. Alternative approaches like MRPT \cite{MRPT2019}, CARMEN \cite{CARMEN2019}, Player \cite{PLAYER2019}, Microsoft RDS \cite{RDS2019}, and others, provide some of the ROS features, but the lack of support for extending across devices imposes major limitations to their development.

The operating basis of our autonomous agent economy would be embodied in a ROS ``behavioral algorithm'' which would enable the interaction of robots with their environment. ROS is composed of many nodes, each providing specific functionality. These nodes communicate with each other by messages, which are themselves data structures. Messages can be passed among nodes by asynchronous (topics) or synchronous (services) mechanisms. Topics are a publish-subscribe method of inter-node communication. When a node publishes a message to one topic, each node that subscribed to that topic will receive that message. On the other hand, services are a request-reply method of inter-node communication. In this case, a node requests a message to another and waits for the reply before continuing \cite{Carvalho2016}.

\begin{figure}
    \centering
        {\includegraphics[width=1.0\linewidth]{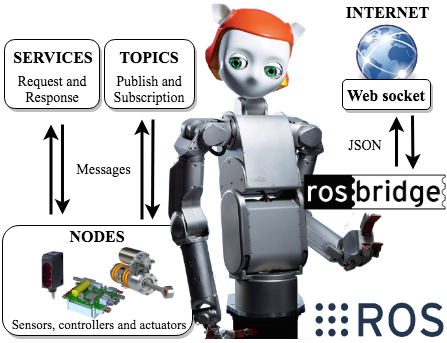}}
\caption{Robot Operating System (ROS) framework. Sensors, controllers and actuators can be interfaced with ROS, allowing communication between them through messages: synchronously via publish/subscription within topics or asynchronously via request/response within services. To communicate with an external network such as the internet the Rosbridge protocol can be used.}
\label{Fig3}
\end{figure}

For intelligent robots to communicate, ROS can also provide communication channels with an external network, in the form of the Rosbridge network interface (Figure \ref{Fig3}). Rosbridge offers simple, socket-based programmatic access for interfacing with web technologies like Javascript. That is, Rosbridge allows access to
underlying ROS messages and services as serialized Java Script Object
Notation (JSON) objects, and in addition, it provides control over ROS
node execution and environmental parameters \cite{Crick2011,Salama2018}.

To conclude, the highest values that the ROS middle-ware offers to a robot economy is the ability to control physical robotic assets and also the implementation and execution of algorithms for robotic systems in a distributed fashion. ROS-based products are already becoming market-ready, addressing   manufacturing, logistics, agriculture, and more. Government agencies are also looking more closely at ROS for use in their fielded systems; e. g., NASA has used ROS for their Robonaut program \cite{Badger2016}. A recent development towards a robot economy using ROS, is called AIRA (Autonomous Intelligent Robot Agent) project \cite{AIRA2017}, and it implements economic interactions between humans and robots or between multiple robots via Ethereum-based smart contracts. 

\subsection{Cloud Robotics and Internet of Robotic Things}

Complex real-world problems, requiring real-time execution, that demand sophisticated data analysis and computational capabilities, are challenging for robots to handle. A potential approach for solving some of the challenges is ``Cloud Robotics'' \cite{Kehoe2015}, enabling ubiquitous, practical, on-demand network access to a shared pool of computing resources (e.g., networks, storage, algorithms, and services).

Cloud robotic architectures consist of two main components: the cloud infrastructure and its physical embodiment, also known as bottom facilities. Bottom facilities include various types of robots while the cloud infrastructure consists of high-performance servers and massive databases, that can support high-speed processing along with huge storage capabilities. The cloud infrastructure cannot only provide the means for a robot which needs external data to support its operation but also opens ways to interact with other robotic systems \cite{Saha2018}. As an example, in 2018 Amazon Inc. launched the AWS RoboMaker service \cite{AMAZON2019}. It does not only ease integration of the ROS framework within Amazon’s cloud-based machine learning services, but it also includes an Integrated Development Environment (IDE) to make coding with ROS easier, and a robot simulation service with pre-built worlds.

The ``Internet of Robotics Things'' (IoRT) \cite{Ray2016} is a relatively new paradigm, which goes beyond networked and collaborative/cloud robotics and integrates heterogeneous intelligent devices into a distributed architecture of platforms operating both in the cloud and on edge computing. IoRT  was originally defined in the above paper as ``intelligent robotic devices that can monitor events, fuse sensor data from a variety of sources, use local and distributed intelligence to determine the best course of action, and then act to control or manipulate objects in the physical world, and in some cases while physically moving through that world''. IoRT arises because multiple autonomous robots must communicate and execute physical tasks in coordination, while exchanging information and granting security \cite{Simoens2018}.

A robot economy relies on multiple autonomous agents exchanging information and coordinating their behavior effectively. Thus, among the technologies that allow the development, implementation and deployment of Cloud Robotics and IoRT applications, the following are particularly relevant for a robot economy: (a) Platform architectures (that is, the application layer that facilitates
communications, distributed computation, data flow and storage, and general applications to be used across robotic platforms to deliver services) that provide versatility, ease-of-use and efficiency; and (b) Communication infrastructure that connect suppliers and services across wide geographic areas and with versatile stakeholder interactions \cite{Vermesan2017}. 

\subsection{Blockchain}

Although the centralized cloud has many advantages, to guarantee the requirements imposed by a robot economy a preferred option is a decentralized and distributed approach. By means of a peer-to-peer periodically updated copy system, the information in a blockchain can be pieced back together in the event of small-scale loss scenarios. Thus, it confers robustness to the economic system with respect to errors in the transmission or storage of data. Moreover, it also guarantees security against digital hackers or data thieves since the data is widely distributed and it can be continuously verified by networks of peers. And, finally, decentralization provides some privacy given that it is not fully controlled or accessible by a third party \cite{DeepCloudAI2019}. Once a viable decentralization has been achieved, ``Blockchain'' technology (originated with the bitcoin cryptocurrency \cite{Swan2015}) for cloud storage and digital transactions starts to provide its full potential to support a robot economy.

The blockchain is a public chronological database of transactions recorded by a network of agents. It is, as its name suggests, the grouping of data sets (referred as blocks) forming a chain. Each block contains: (a) information about a certain number of transactions (individual transactions containing details of who sent what to whom); (b) a reference to the preceding block in the blockchain; and (c) an answer to a complex mathematical challenge known as the ``proof of work''. The proof of work is used to validate the data associated with that particular block, as well as to make the creation of blocks computationally ``hard'', thereby preventing attackers. After ensuring that all new transactions to be included in the block are valid (and do not invalidate previous transactions), a new block is added to the end of the blockchain by an agent (referred to as a the ``miner'') in the network. At that moment, the information contained in the blockchain can no longer be deleted or modified, and it is available to be certified by everyone in the network \cite{Castello2017}. Therefore, the blockchain can be considered as an open and distributed ledger that can record transactions between two parties in a verifiable and permanent way. 
The blockchain technology allows, by combining peer-to-peer networks with cryptographic algorithms, for a group of autonomous agents to: (1) reach an agreement on a particular affair; and (2) record that agreement without the need of a controlling authority. The blockchain is also a payment mechanism and makes it possible for autonomous agents to exchange goods and services among themselves using crypto currencies \cite{Djukic2018}.

Within the blockchain context, smart contracts mean transactions that go beyond simple buy-sell transactions, and may have
more extensive digital instructions embedded into them. A smart contract may act as an autonomous entity on the blockchain. It has its own digital signature on the blockchain and it is both defined by the blockchain code and potentially controlled by the code itself \cite{Hanada2018}. Three elements in smart contracts that makes them distinct are certain levels of autonomy, self-sufficiency, and decentralization \cite{Swan2015}. 

A blockchain smart contract between a supplier and a buyer provides the trust that otherwise would be required during human auditing processes. Not only can blockchain contracts contain the same level of detail as traditional contracts, but they can do something no conventional contract can do: to automatically perform tasks such as negotiating prices and self monitoring inventory \cite{Cognizant2016}. 

\begin{figure}[t]
    \centering
        {\includegraphics[width=1.0\linewidth]{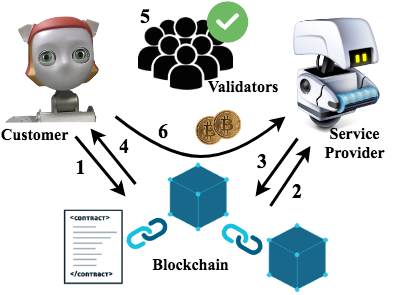}}
\protect\caption{Our proposed interaction between autonomous robots via smart contracts: (1) A robot customer creates a smart contract; (2) Another robot, that provides services, executes the contract; (3) After the contract is finished, a response is sent to the blockchain; (4) The customer receives the response; (5) Blockchain peers, validate if everything is correct before approving the transaction; (6) A payment is executed.}
\label{Fig4}
\end{figure}

We propose a high-level scheme for robot-to-robot economic interactions via smart contracts as shown in Figure \ref{Fig4}. As a first step, a smart contract in the blockchain and corresponding instructions for a requested service by a robot customer are sent out to the blockchain in search of a service provider. A service robot may accept the request of the service and performs the task. Once finished, a service response is sent back to the robot customer, who creates a currency transfer request in the blockchain. A notification of this transfer request is sent to blockchain peers that are prompted to check for correctness of the service and transactions. The liability of the transaction is then recorded in the blockchain. Finally the robot service provider is paid by the robot customer through the blockchain \cite{Danilov2018}. 

Thus, with the integration of Blockchain, the ROS interface for physical systems, Cloud Robotics, and IoRT, autonomous robotic agents could find a starting point for the technology needed in a robot economy.

\section{Socio-economic impact of a robot economy}

We believe that we are in a phase in which robots will have an impact on labor dynamics and in the broader socio-economic context. The technological evolution is evident in view of the current growth of robotic efforts aimed at developing abilities for robots to perform physical and cognitive work. 
 
Currently, service robots are being provided by well established robotic companies such as Kiva, KUKA, Fanuc, iRobot, Toyota, Amazon, and more. Our proposed technology will require robots to operate in unstructured environments while interacting with humans. Several companies already commercialize collaborative robots (cobots), such as Universal Robots, Boston Dynamics, Waymo, MiR, among others. Several startups are exploring more daunting applications, involving vision guided manipulation, learning from demonstration, and natural language interactions, such as Energid, Thinking Robots, Dextery, and Diligent Robotics. Other robotic companies are focusing on semiautonomous interactions with human drivers, such as Oshkosh Corp and DCS Corp. Finally, there are many new companies offering robots as a service (RaaS) including cleaning, security, delivery, and warehouse robots. Some examples are Maidbot, Knightscope, Kiwibot, Kindred, and Fetch Robotics. Although several of these companies offer requesting robot services via cloud applications, none provide a framework for autonomous robot-to-robot services.

As new robotic technologies are developed questions arise: Which jobs are vulnerable? What will be the reaction of our institutions? What is the outcome of a robotic economy with respect to global growth and income distribution? In the short-term, could a robot economy result in negative social outcomes? Our institutions are expected to react to rapid changes on employment. There is an ongoing discussion according to some macroeconomic models on whether an unregulated society could face challenges in employment \cite{Sachs2015,Berg2018,Stiglitz2018}. Due to increased technological productivity, the loss of jobs need to be balanced with the creation of new ones while providing institutional means to increase the training of our labor forces and economically support people at risk \cite{Ivanov2017a}.

Occupations that share a predictable pattern of activities will slowly be replaced by robots. Jobs that require creativity, complex judgment and lack of structure will remain unscathed. Labor for low-skill workers will slowly be transformed into more creative and social oriented jobs which our populations will adapt to over generations given our institutional efficacy. Some people believe that a robot economy could produce imbalances in the current distribution of income that must be corrected to maintain a social equilibrium \cite{Berg2018,Freeman2015,Stiglitz2018}. From their perspective, the distribution of income seems to shift towards businesses that have more robots. Therefore, growth might be higher for those who are wealthier and able to invest in robotics, increasing the gap with those who live paycheck-to-paycheck \cite{Freeman2015}. So, it will be logical to implement social measures to ensure that the returns from robotic assets benefit the wider populations. According to \cite{Berg2016,Freeman2015}: ``workers need to own part of the capital stock that substitutes for them to benefit from these new robotic technologies: workers could own shares of the firm, hold stock options, or be paid in part from the profits''.

The positive outcomes of robotics expected from the effectiveness of our institutions can create great growth beyond traditional solutions. A recent McKinsey report \cite{McKinsey2017} estimated that automation could raise productivity growth on a global basis by as much as $0.8$ to $1.4$ percent annually. Another recent PwC report \cite{PwC2018} estimated that ``global GDP in 2030 could be significantly higher than that of 2016 as a result of the economic impact of intelligent robotics. This impact will be driven by (a) productivity gains from process and industrial automation as well as providing AI technologies to the labor force (assisted, autonomous and augmented intelligence); and (b) an increase in the consumer demand resulting from the availability of higher quality products and services''.

In the long-term, intelligent robotics could overcome the physical limitations of capital and labor and represent a new source of value and growth. There seems to be a degree of agreement on the benefits of this outcome, for example, the European Robotics Research Agenda 2020 outlines current developments in the following way: ``The robotics technology will become dominant in the coming decade. It will influence every aspect of work and home. Robotics has the potential to transform lives and work practices, raise efficiency and safety levels, provide enhanced levels of service and create jobs. Its impact will grow over time as will the interaction between robots and people'' \cite{euRobotics2014}.

Therefore, a robot economy might have a large positive impact in our society and our economy and we can help our institutions to overcome the barriers. As the 2016 White House report called ``Artificial Intelligence, Automation, and the Economy'' points out: ``with the appropriate attention and the right policy and institutional responses, advanced automation can be compatible with productivity, high levels of employment, and shared prosperity'' \cite{WhiteHouse2016}.

\begin{figure}[t]
    \centering
        {\includegraphics[width=1.0\linewidth]{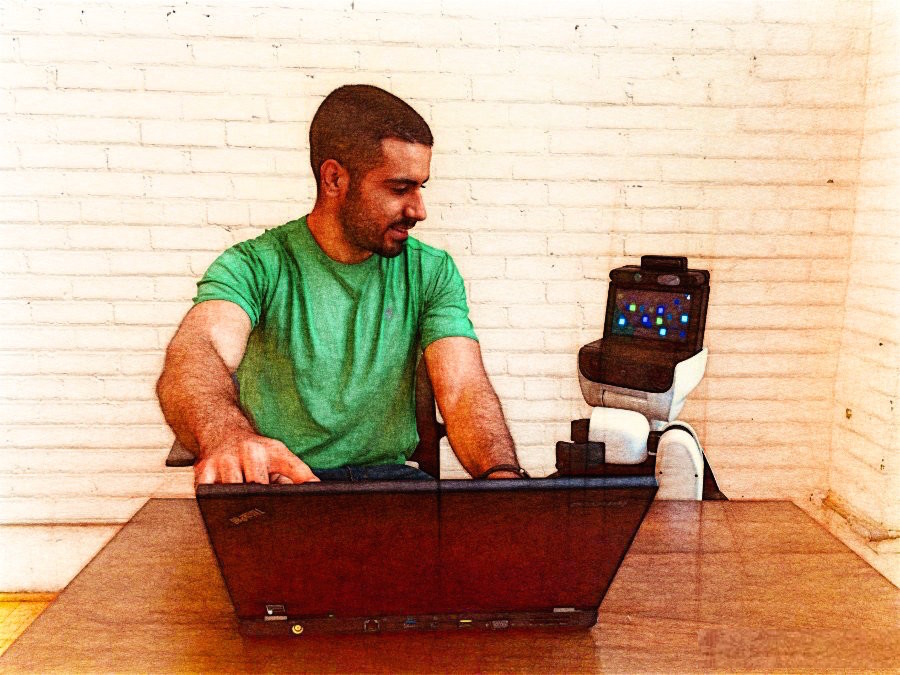}}
\protect\caption{Intelligent robots have the potential to improve people's well-being and to provide new sources of value and growth.}

\label{Fig5}
\end{figure}

\section{The cleaner robot model}

\begin{figure*}[t]
    \centering
        {\includegraphics[width=1.0\linewidth]{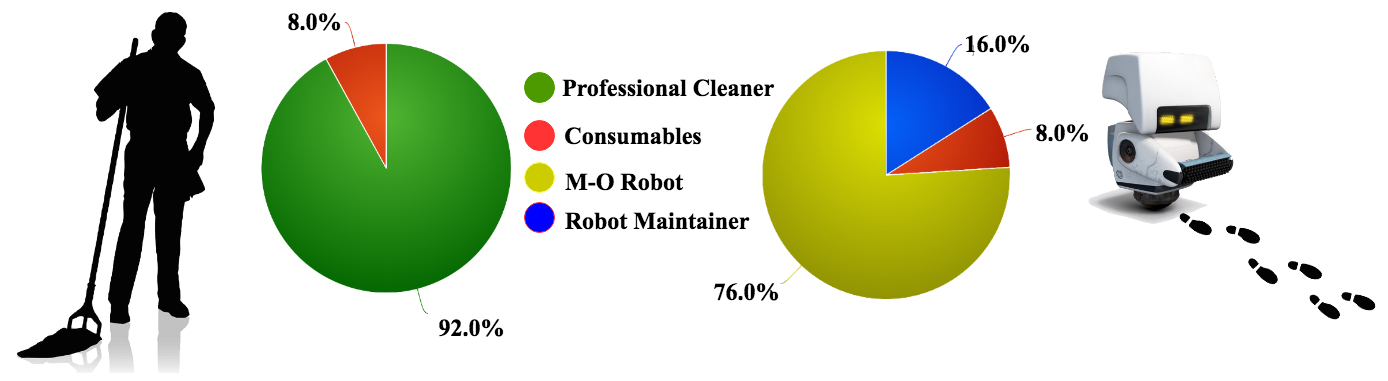}}
\caption{Distribution of Maurice's budget for keeping a public hall clean. On the left side it is shown the cost share when a professional cleaner is hired. On the right side it is shown the cost share when an M-O robot is ``hired''.}
\label{Fig6}
\end{figure*}

Modeling the impact of a robot economy at a large scale involves many complex social questions that cannot be incorporated into a simple macroeconomic model. However, in a qualitative way, we can show how
the concepts mentioned in the previous sections are applied in a real
situation through a simplified model. The example we developed here has been adapted from an economic study that is described in \cite{Prassler2008}. Maurice is passionate about robotics and through his job has to ensure the cleanliness of a $600\,m^{2}$ entry hall of a public building. Let us briefly look into professional cleaning services with approximate figures that serve as estimates for our case study but are not directly applicable to a specific application case.

The professional cleaning of $1\,m^{2}$ floor costs on the order of $\$\,0,10$ if done manually. This includes the cost of labor and
materials. If the $600\,m^{2}$ entry hall of a public building is cleaned once a day, five days a week, $52$ weeks a year, the total
cost of cleaning is $\$\,15600$ per year, of which approximately $\$\,1200$ correspond to consumables and the rest to labor.

Suppose a new state-of-the-art cleaning robot called M-O appears in the market as a service. That is M-O is a particular robot and not a type of robot. M-O's functionality is enabled by a novel architecture that combines ROS, IoRT and Blockchain technologies. Therefore, the M-O robot has the ability to offer its services through smart contracts while working as an autonomous agent. As an amateur roboticist, Maurice decides to hire the services of M-O to perform the cleaning task, replacing the services of a human cleaner. Maurice prepares and processes through a public blockchain a smart contract with M-O, for providing its services. The contract defines the tasks that M-O must perform and the agreement for the payments that Maurice will make for the services contracted. Consequently, M-O will be in charge of the cleaning of the entry hall, and replacing the cleaning materials and the necessary work for its own maintenance.

Let us further assume that deploying and maintaining the cleaning robot requires the work of a human specialist for $20\,\min$ per day and that the specialist has an hourly wage of $\$\,30$. Thus, the labor for deploying and maintaining the cleaning robot costs
approximately $\$\,2600$ per year. Let us assume, also, that consumables and repair add up to $\$\,1200$ per year. Finally, assuming a depreciation period of four years and, thus, considering that the annual amortization
cost of the cleaning robot is $\$\,11800$ (the price for a professional cleaning robot is around $\$\,45000$), the annual cost of robotic cleaning results approximately in $\$\,15600$.

That is, with these assumptions, the total cost to Maurice for both type of services is the same, whether the cleaning is done manually or performed with a robot. In the first case more than $92\%$ of the total expenditure budget corresponds to the cost of a worker, which we can assume that,
in turn, will be used by the worker for expenses of food, housing, clothing, social security, transportation, etc (Figure \ref{Fig6}). However, in the second case, less than $16\%$ of the expenditure budget is allocated to the direct remuneration of a worker, while $76\%$ of the said budget corresponds to the remuneration of the capital invested in services regarding the cleaning robot (amortization, capital insurance, etc.).

Some consequences of the potential impact of a robot economy could be analyzed from this example: First, it is evident that there has been an effective displacement of a human cleaner since it has been replaced by the robot; Second, given that the human cleaner is not immediately relocated, at least in the short term, an increase in unemployment might occur; Third, although the usage of the cleaning robot has displaced the human cleaner, a new high-skilled labor activity such as the robot maintenance has appeared; Fourth, the cleaning robot has changed the distribution of the cost budget involving a transfer of labor costs towards expenses in capital retribution, shifting the cost towards the capital invested in the robot; Fifth, it should be noted that there is a net contribution to growth since the worker who performed the manual cleaning is now available to perform other tasks, increasing the potential for high-skilled production growth; Finally, the capital source that invests in the robot ``earns'' a significant part of the income, so that the question of ``who owns the robots'' acquires great relevance in the final distribution of income.

This example aims to illustrate the effect of incorporating service robots in our economy. Right now, service robots have potential market use in several new applications across sectors, including logistics, education, healthcare, domestic tasks, personal and elderly assistance, home security, agriculture and entertainment and more. Exoskeletons, or human-robot hybrids, are robots connected to the human body whose primary field of application is rehabilitation, but which are also designed as an aid workers to support heavy-duty processes. Those systems could also participate in a robot economy system.

Several service robots are expected to hit the market with advanced capabilities to assist humans of different age groups from children to the elderly. At the Consumer Electronics Show (CES) 2018, Aeolus Robotics launched a multifunctional in-home robot powered by artificial intelligence. Also at CES 2018, Honda introduced its new 3E (Empower, Experience, Empathy) Robotics Concept, designed to assist people in various tasks, learning from their interactions with people to improve their operation and become more empathetic.

It is expected that the service robotics market expansion will be favored by a higher return on investment with a shorter payback period. In fact, blockchain platforms that intend to facilitate service robot operations are already being developed. As an example, the Robotou ecosystem  \cite{ROBOTOU2019} is a centralized blockchain-enabled platform, supplying various robotic services to address labor force shortages, and improve human-robot user experience. This platform offers individuals to become owners of robots, redefining the ``buy-to-let'' concept in the age of AI. To achieve this goal, Robotou will establish transparent and reliable governance mechanisms in the robot service ecosystem, adopt encryption protocols to protect valuable data and intellectual properties, adopt smart contracts to improve efficiency, and utilize tokens to reward good behaviors within the ecosystem. Robotou already has validated use cases of catering robots in Hong Kong and Beijing restaurant chains, in which every robot will run about 100 times each day, delivering 4-6 dishes per run \cite{ROBOTOU2019}. The objective of this platform is to initiate a trend where robotic services are integrated in our everyday lives, creating alternative income sources for robot and business owners.

\section{Robot economy challenges}

Robot economies could be right around the corner. With this inflection point approaching, it is important to contemplate all aspects related to legal frameworks, security, ethics and the future of robotics. Regulation must consider the social effects without stifling innovation
\cite{EuParl2017}. The development of such economies creates several
regulatory opportunities. Firstly, an important concern arises if intelligent robots operate without the intervention of humans and, more acutely, when they do it without people's awareness. Concerns must be raised about the liability of robots, norms and privileges for robots, safety, security, and privacy to ensure fairness and continued progress of human societies despite their growth \cite{Alexandre2017}. Secondly, there is no global consensus on authorizing cryptocurrencies as an international payment method, due to the lack of control with existing monetary policies and concerns over criminal exploitation. We look forward to see what opportunities appear in the future. And finally, robots do not currently have ``legal personhood'' to engage into contracts. There have been initiatives, e.g. \cite{EuParl2017}, based on the idea of creating an electronic personhood for robots, distantly comparable to the legal personhood available for business organizations, but these proposals have been so far controversial, since they raise conflicts related to liability\cite{Caytas2017,EuP2016}.

Another key concern about a robot economy is global and individual security. A robot assistant may need access to databases of all kinds, as well as personal information to create complex models to effectively adapt to people's needs given actual circumstances. Acquiring this kind of data requires collecting information from environmental and human activity sensors taken during our daily lives including physical, cognitive and emotional states. This kind of information can be extremely sensitive and could compromise individuals' dignity and our right for privacy. Deploying these technologies at home or business place, for instance, for marketing and product development purposes should be subject to an increase of regulatory mechanisms to provide citizens and governments with the ability to counteract negative effects and fair use of these technologies. 

The robot economy is gaining pace, but is it possible that it contradicts our values or displaces us \cite{Yang2018}? If this was the case, our human condition could worsen and we could lose personal and economic freedom. However, we believe that advances in intelligent robots will be inevitably controlled by human institutions which will prevent them from competing against us. These kind of key questions are studied by influential researchers such as Carme Torras \cite{Torras2018}.

The progressive development of a robot economy is expected to effectively increase people's well-being. We encourage that the economic and social objectives of robotics and AI research are geared towards direct improvements to our societies \cite{EU-OSHA2015}. The main challenge of robotization lies in effectively interfacing human and robot capabilities. The advantage of robotics lies on their ability to increase productivity without increasing costs, or even decreasing costs, i.e. ``blue-ocean'' technologies. Humans alone are limited in their productivity due to physical fatigue and our ability to systematically process batch data and goods. Thus robots seem to be deemed to occupy markets that require batch production, intense physical tasks and augmenting human cognitive capabilities. There is no widespread interest or evidence on a global replacement of human capabilities. 

Finally, if we want to make the robot economy a key factor of prosperity, one interesting question that arises is: ``Who will own the robots?''
\cite{Berg2016,Freeman2015,Rotman2015}.

\section{Conclusions}

This paper discusses a framework for a robot economy with currently available technologies, and reviews some challenges it faces in the current socio-economic scenario. A robot economy is described as the economic system that uses physical robots to greatly increase production in addition to human labor and in which intelligent robots can perform some economic operations not allowed or possible today.

There is current technology that would allow robots to perform on-board tasks related to economic activity and peer-to-peer communications. The blockchain technology, as a peer-to-peer payment system and as a digital currency, allows robots to execute programs specified in terms of intelligent contracts, creating a form of generally accepted compensation for their work. In addition, blockchain can be useful to register robot transactions and as a means of payment. Thus, with the integration of blockchain, network and control middle-ware, a framework for a simplified robot economy has been discussed here.  

With our framework, intelligent robots will have great flexibility of operation, increased demand, and certain contractual freedom. However, it is clear that not all the mentioned activities can be fully achieved on a contractual basis, and, in addition, ultimately the contractual framework must be approved by humans, so the autonomy of intelligent robots cannot be complete. But we believe that a high degree of autonomy can be reached so it incentivizes the robot economy.

Our current global and local institutions seem to be effective for self-regulating the negative effects of evolving technologies such as robotics and AI. Therefore, without entering in detail we don't expect to see radical and quick changes on governance but a natural evolution of our institutions to self-regulate this technology as time goes by. Robots are great good and service producers but they are not consumers, unlike humans. There is no reason to expect that robots would replace humans as they would be required to develop human-like needs which seems unlikely. Speculations on this issue seems to be out of context as a topic for this paper. Therefore, we look forward to see institutions creating mechanisms for fair use of robotics and AI. 

Several complex questions regarding legislation, security, ethics and the future of robotics will arise while robots enter our economies in more depth. Intelligent robotics offer an unimaginable spectrum of possibilities, and it is roboticists' responsibility to get educated on economic and institutional effects to make correct statements on the consequence of economic automation.

Finally, we don't believe there is a threat with intelligent robots for the reasons above. On the contrary, robots will increase productivity and benefit us all as all previous historic technology waves have benefited human societies. It is crucial to create awareness of the reality of a robot economy, because ready or not, here it comes.

\section*{Acknowledgments}
The authors would like to thank Carme Torras from Institut de Robòtica i Informàtica Industrial (CSIC-UPC) in Barcelona, Spain. 

\bibliographystyle{spphys}
\bibliography{Referencias}

\end{document}